\documentclass[letterpaper, 10 pt, conference]{Styles/ieeeconf}  
\IEEEoverridecommandlockouts                              
\let\labelindent\relax
\usepackage{cite}
\usepackage{graphicx}
\usepackage{graphics}
\usepackage[caption=false]{subfig}
\usepackage{mathtools}
\usepackage{amsmath}
\usepackage{amsfonts}
\usepackage{makecell}
\usepackage{changepage}
\usepackage{dirtytalk}
\usepackage{pgfplots}
\usepackage[font=small]{caption}
\usepackage{enumitem}
\usepackage{multirow}
\usepackage{makecell}

\title{\LARGE \bf
ModGNN: Expert Policy Approximation in Multi-Agent Systems with a Modular Graph Neural Network Architecture
}

\author{Ryan Kortvelesy and Amanda Prorok  
\thanks{Ryan Kortvelesy and Amanda Prorok are associated with the Department of Computer Science and
Technology, University of Cambridge, Cambridge, United Kingdom (Emails: rk627@cam.ac.uk, asp45@cam.ac.uk)}
}

\begin{document}

\maketitle
\thispagestyle{empty}
\pagestyle{empty}

\begin{abstract}
Recent work in the multi-agent domain has shown the promise of Graph Neural Networks (GNNs) to learn complex coordination strategies. However, most current approaches use minor variants of a Graph Convolutional Network (GCN), which applies a convolution to the communication graph formed by the multi-agent system. In this paper, we investigate whether the performance and generalization of GCNs can be improved upon. We introduce ModGNN, a decentralized framework which serves as a generalization of GCNs, providing more flexibility. To test our hypothesis, we evaluate an implementation of ModGNN against several baselines in the multi-agent flocking problem. We perform an ablation analysis to show that the most important component of our framework is one that does not exist in a GCN. By varying the number of agents, we also demonstrate that an application-agnostic implementation of ModGNN possesses an improved ability to generalize to new environments.

\end{abstract}

\section{INTRODUCTION}
Graph Neural Networks (GNNs) \cite{OriginalGNN, GraphNetworks} are valued for their ability to find relationships in data that exhibit an underlying graph structure. They are able to generalize to new data and new graph structures by learning local operations \cite{Ribeiro}. In the past, GNNs have been used for countless tasks, including node classification \cite{NodeClassification, GraphAttention, GraphSAGE}, graph classification \cite{GraphClassification}, and link prediction \cite{LinkPrediction}. More recently, they have been applied to multi-agent problems \cite{ArbaazLabelled, ArbaazUnlabelled, TolstayaFlocking, QingbiaoMAPF, GNNCoverage, ProrokWorkshop}.

While there is a diverse collection of GNN architectures for various applications, development in the field of multi-agent systems is still in its incipient stages. Most of the current applications \cite{ArbaazLabelled,ArbaazUnlabelled, TolstayaFlocking,QingbiaoMAPF} use graph convolutional networks (GCNs), which apply a graph convolution followed by a nonlinearity \cite{GamaGCN}. This represents a slight departure from the theoretical basis of pure graph convolutions, but it has been shown that adding a nonlinearity boosts the model's performance, allowing it to emulate many different policies \cite{Ribeiro}.

This raises a few questions: If we add more nonlinear models in between the aggregation steps, will that expand the set of policies which the GNN can imitate? Will it improve generalization? Would a learnable network be more effective before or after each of the aggregation steps?

In this paper, we seek to answer these questions by creating a decentralized general GNN framework for multi-agent applications which serves as an extension of GCNs. We identify the operations in a GNN which are absolutely essential, and fill the gaps in the resulting architecture with user-defined submodules. In our implementation of this framework, we define these functions as multi-layer perceptrons (MLPs) to approximate arbitrary nonlinear functions.

We use the flocking problem to evaluate this implementation of our framework against multiple baselines, including two variants of the GCN. The comparison also includes several implementations of our framework with individual submodules removed, which serves as an ablation analysis. We compare the performance of these models trained on different values of $K$ (the number of communication hops). Furthermore, we test the generalization of the models by evaluating their ability to handle a different number of agents than they saw during training.

In this paper, our main contributions are:
\begin{itemize}
    \item The development of the ModGNN framework, which not only generalizes the architectures of existing models as a basis for comparison, but also identifies \say{submodules} which are rarely used in existing architectures\footnote{The ModGNN Python library is available at https://github.com/proroklab/ModGNN}.
    \item A theoretical analysis of the benefits of those rarely-used submodules within the ModGNN framework, as well as experimental results to back up that analysis.
    \item An experimental comparison of the expressivity of existing models and a naive implementation of ModGNN. This comparison is used to evaluate the relative performance of each submodule in the ModGNN framework.
\end{itemize}

\section{PRELIMINARIES}

In this section, we formalize the problem and provide a formulation for GCNs, the architecture that we use as a baseline.

\subsection{Problem Formalization}

In a multi-agent problem with $N$ agents, we consider the communication graph formed by the agents acting as nodes, and communication links between those agents acting as edges. If agent $j$ is within communication range $R_\mathrm{com}$ of agent $i$ at time $t$, then we say that agent $j$ is in agent $i$'s neighborhood $j \in \mathcal{N}_i(t)$. Furthermore, we define the $k$-hop neighborhood of agent $i$ as the set of agents $j$ such that a walk through the communication graph of length $k$ from agent $i$ to agent $j$ exists. For example, the $0$-hop neighborhood of agent $i$ is the singleton $\{i\}$, and the $1$-hop neighborhood is the set of its neighbors $\mathcal{N}_i(t)$.

In ModGNN, we select a value $K$ that defines the maximum number of hops that data is permitted to travel. Consequently, the available data at each node is the set of aggregated data from each $k$-hop neighborhood from $0$ to $K$.

When implementing ModGNN, one must also select the number of layers $L$. For agent $i$, the output of layer $l$ and input to layer $l+1$ is denoted $\mathbf{x}_i^{(l)}(t)$. The input of the first layer $\mathbf{x}_i^{(0)}(t)$ is defined as the raw observation $\mathbf{o}_i(t)$. The output of the last layer $\mathbf{x}_i^{(L)}(t)$ is the action $\mathbf{u}_i(t)$.

\subsection{Graph Convolutional Networks}
\label{section:GCN}

It is simplest to define a GCN from a centralized point of view \cite{GamaGCN}. First, the aggregated information from a $k$-hop is collected with the expression $\mathbf{S}^k \mathbf{X}^{(l)}$, where $\mathbf{S} \in \mathbb{R}^{N \times N}$ is a graph shift operator (GSO) such as the adjacency matrix, and $\mathbf{X}^{(l)} = \left[\mathbf{x}_1^{(l)}, \ldots, \mathbf{x}_N^{(l)} \right]$ is the stacked data at layer $l$ of all agents. The output $\mathbf{X}^{(l+1)}$ of the GCN is computed by applying a graph convolution and point-wise nonlinearity

\begin{equation}
    \mathbf{X}^{(l+1)} = \sigma \left( \sum_{k=0}^K \mathbf{S}^k \mathbf{X}^{(l)} \mathbf{A}_k  \right)
\end{equation}

where each $\mathbf{A}_k$ is a filter tap in the graph convolution filter. If $\mathbf{X}^{(l)}$ has dimension $N \times D_l$ and $\mathbf{X}^{(l+1)}$ has dimension $N \times D_{l+1}$, then each $\mathbf{A}_k$ has dimension $D_l \times D_{l+1}$. The input to layer $1$ for agent $i$ is defined as the raw observation $\mathbf{x}^{(0)}_i := \mathbf{o}_i$, and the output of layer $L$ for agent $i$ is defined as the action $\mathbf{x}^{(L)}_i := \mathbf{u}_i$.

In Section \ref{section:Models}, we demonstrate how the same GCN architecture can be formulated in the context of the ModGNN framework.

\section{MODEL ARCHITECTURE}
\label{section:ModelArchitecture}

The ModGNN framework is developed from the ground up, starting with a few basic assumptions. First, we assume that there is a bound on the amount of data that can be transmitted. That is, an agent cannot receive all of the raw data from its $K$-hop neighborhood. Therefore, there must exist some form of aggregation in the communication stage of the GNN. A framework without any aggregation in the communication stage might be slightly more flexible, but it would also cause the volume of transmitted data to be exponential in $K$, which is clearly not scalable.
Second, we assume that the most logical grouping for the data is by $k$-hop neighborhood. One reason for this is that many models (such as GCNs) group by neighborhood, so this allows our framework to serve as a generalization of those models. Perhaps a more compelling reason is that this scheme allows the model to keep the data as \say{separate} as possible, and thus preserves information. For example, the prevailing alternative method is to aggregate all data, updating a recurrent state at each node \cite{ProrokWorkshop, GNNCoverage}. By performing more aggregation than our proposed method, it becomes impossible to obtain individual measurements for distinct neighborhoods, and thus information is lost.
Lastly, we assume that the best aggregation operator is summation. Again, this is partially motivated by the fact that most GNN models use summation, so we can allow ModGNN to serve as a generalization of those models by using it. Additionally, summation possesses a few desirable properties---it is permutation invariant, and unlike operations like $\max$, it combines information from all of its inputs.

\begin{figure}[]
      \centering
      \includegraphics[width=0.45\textwidth]{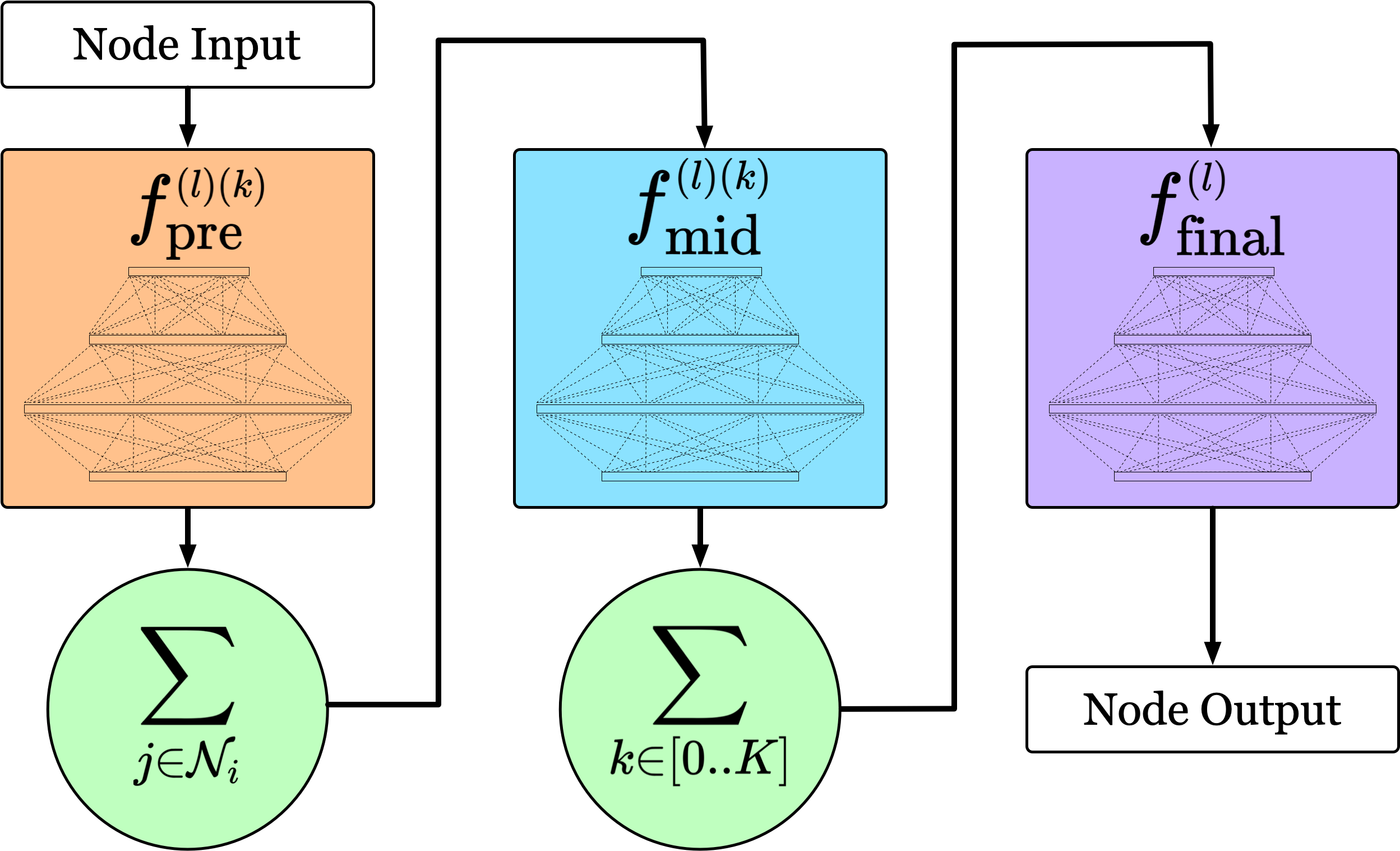}
      \caption{ModGNN's node update module. First, $f_\mathrm{pre}$ is applied to the incoming transmissions from each neighbor, and the data from those neighbors is aggregated together. Next, the combined data from each of the $K+1$ neighborhoods is passed through $f_\mathrm{mid}$ and aggregated together. Lastly, $f_\mathrm{final}$ is applied to produce the final output.}
      \label{fig:GNNNode}
\vspace*{-8pt}
\end{figure}

These three assumptions generate a skeleton structure for a GNN by defining exactly where the aggregation operations take place (Fig. \ref{fig:CommunicationSystem} and Fig. \ref{fig:GNNNode}). The summations in the message aggregation module (Fig. \ref{fig:CommunicationSystem}) and the first summation in the node update module (Fig. \ref{fig:GNNNode}) combine data within the same $k$-hop neighborhood, and the second summation in the node update module (Fig. \ref{fig:GNNNode}) combines data from different neighborhoods into a single output vector.

Given this skeleton structure, ModGNN defines the most general possible framework by placing user-defined submodules between every pair of aggregations: $f_\mathrm{input}$, $f_\mathrm{com}$, $f_\mathrm{pre}$, $f_\mathrm{mid}$, and $f_\mathrm{final}$. Since most existing GNN models can be implemented in the ModGNN framework, their differences can be analyzed by comparing their implementations for each of the submodules. For example, the differences between standard, attention-based, and message-passing GNNs can be defined entirely within the implementation of $f_\mathrm{com}$.

The most exciting consequence of comparing various GNN models within the context of the ModGNN framework is that an often-overlooked submodule is exposed: $f_\mathrm{pre}$. Most architectures (including GCNs) aggregate the incoming transmissions from each node's set of neighbors \textit{before} that node can start processing the data. On the other hand, the ModGNN formulation offers an opportunity to process the data from each individual neighboring agent. This provides theoretical benefits (discussed in Section \ref{section:GNNnode}) which are reflected in our results (Section \ref{section:Results}).

\subsection{Multi-Layer Architecture}
\label{section:MultiLayerArchitecture}

\begin{figure}[]
      \centering
      \includegraphics[width=0.45\textwidth]{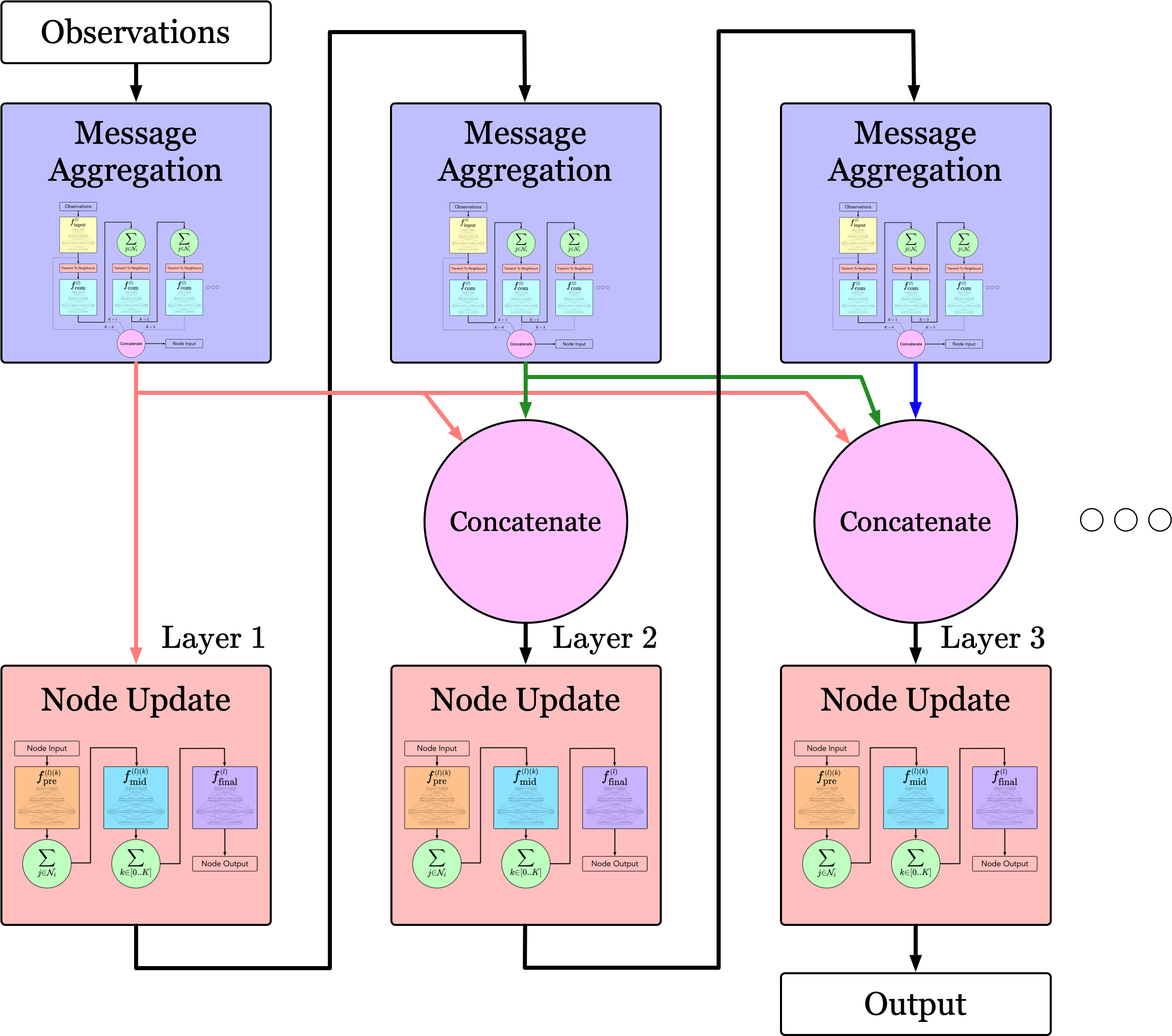}
      \caption{The entire multi-layer architecture of ModGNN. At each layer, the message aggregation module disseminates the output from the last layer, and then the node update module uses the data from all of the previous layers to compute an output.}
      \label{fig:WholeArchitecture}
\vspace*{-8pt}
\end{figure}

The ModGNN framework provides infrastructure for multi-layer GNNs. Each layer consists of a message aggregation module to transmit data, and a node update module to compute the output (Fig. \ref{fig:WholeArchitecture}). For extra flexibility, the input consists of the outputs of all previous layers, but this extra data can easily be masked out if it is not required. Each layer can have different architectures for analogous submodules, but in some cases it makes sense to use the same architecture with parameter sharing.

\subsection{Message Aggregation Module}
\label{section:MessageAggregationModule}

\begin{figure}[]
      \centering
      \includegraphics[width=0.45\textwidth]{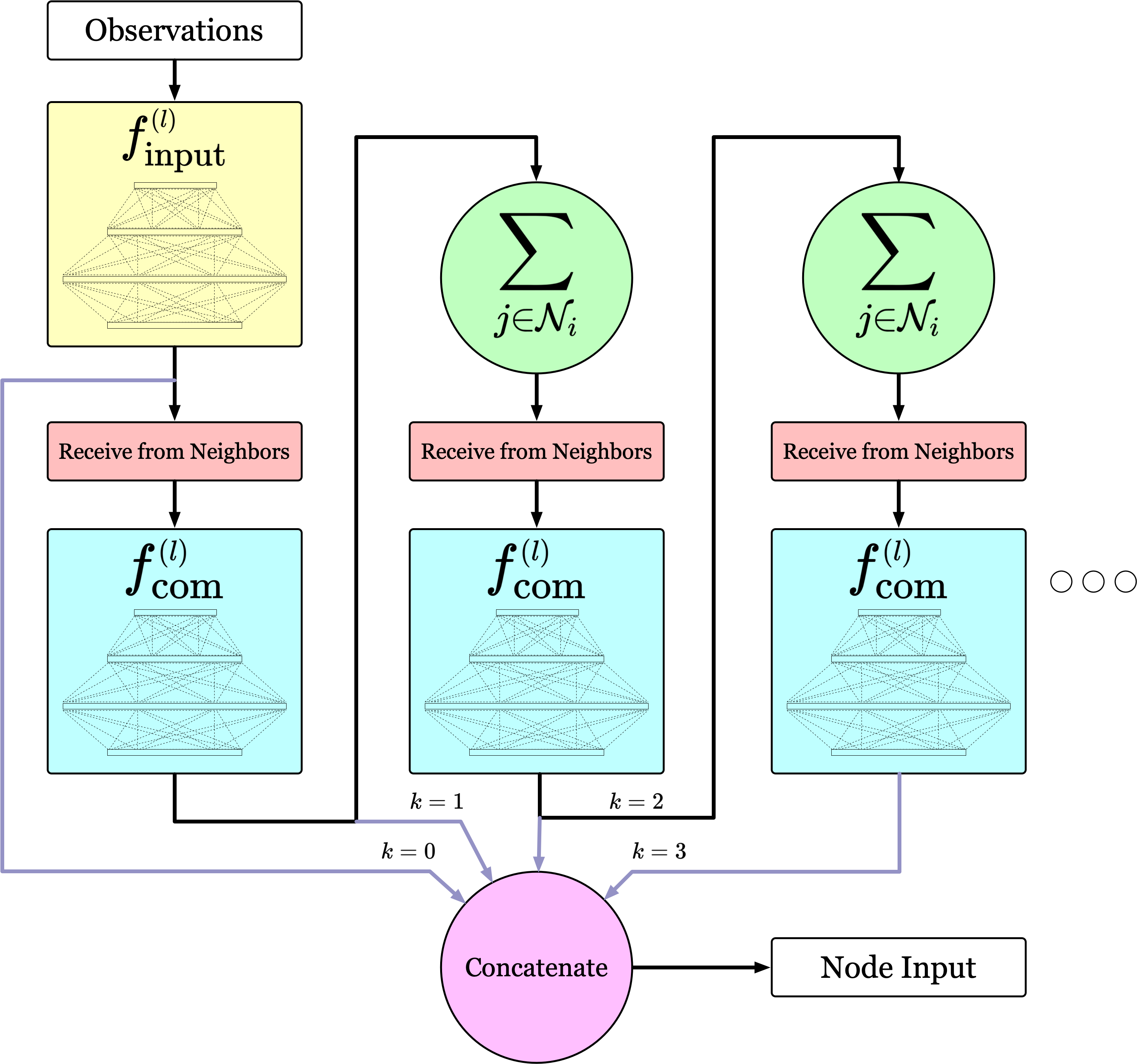}
      \caption{ModGNN's message aggregation module. In this diagram, the system is shown from a centralized point of view. First, the raw observation or output from the last layer is transformed by $f_\mathrm{input}$. Then, for each transmission up to $K$ hops, the data from the neighboring agents is passed through $f_\mathrm{com}$ and then aggregated. The output is the set of the data from each $k$-hop neighborhood up to $K$.}
      \label{fig:CommunicationSystem}
\vspace*{-8pt}
\end{figure}

The message aggregation module handles all communication between agents. Its inputs are the local observations and a transmission from each of the neighbors.

The first step in our message aggregation module is to compress the raw observation with an $f_\mathrm{input}$ function (Fig. \ref{fig:CommunicationSystem}). This step is not only useful for transforming the observation into the desired shape (for example, a CNN can be applied to image observations to flatten the data before it is transmitted \cite{QingbiaoMAPF})---it also provides an opportunity for the observation to be transformed before all of the aggregation steps. Aggregation is a lossy operation, so it is important to transform the data into a space that preserves the most important information. Qualitatively, the purpose of the $f_\mathrm{input}$ submodule can be viewed as learning which information to communicate.

We define $\mathbf{c}_i^{(l)}(t)$ as the compressed observation of agent $i$ at layer $l$. This is calculated by applying $f_\mathrm{input}^{(l)}$ to the set of outputs from all of the previous layers:

\begin{equation}
\label{eq:finput}
\mathbf{c}_i^{(l)}(t) = f_\mathrm{input}^{(l)}\left(\left\{\mathbf{x}_i^{(m)}(t) \; \middle| \; m \in [0..l-1] \right\}\right) .
\end{equation}

The next step is to transmit data from each agent to all of its neighbors. The data from previous timesteps are cached, so an agent can obtain $k$-hop data at time $t$ by requesting $(k-1)$-hop data from time $t-1$ from its neighbors. The benefit to this communication scheme is that only one communication is required per timestep. The GCN does not specifically define a message aggregation module because it is formulated in a centralized setting, but the naive method is to perform $K$ successive communication steps. Consequently, assuming that the evaluation time of the model is negligible compared to the time it takes for perception and communication, ModGNN is able to run K times faster than a naive communication system.

Every time agent $i$ receives data from its neighbors, the $|\mathcal{N}_i(t)|$ incoming vectors are passed through an $f_\mathrm{com}^{(l)}$ function, and then aggregated together (Fig. \ref{fig:CommunicationSystem}). The $f_\mathrm{com}^{(l)}$ submodule defines how the data is transformed as it is communicated between agents. For example, if $f_\mathrm{com}^{(l)}$ subtracts the local state from each incoming state, then it is equivalent to using the Laplacian as a graph shift operator. One can also use $f_\mathrm{com}^{(l)}$ to implement an attention mechanism \cite{GraphAttention} or a coordinate transformation system to shift the observations into the local reference frame.

Let $\mathbf{y}_{ij}^{(l)(k)}(t)$ be the data in layer $l$ from a $k$-hop neighborhood received by agent $i$ from agent $j$ at time $t$. We define $\mathcal{Y}_i^{(l)(k)}(t)$ as the set of all transmissions that agent $i$ receives at time $t$ from a $k$-hop neighborhood in layer $l$:

\begin{equation}
\label{eq:Yk}
\mathcal{Y}_i^{(l)(k)}(t) = \left\{ \mathbf{y}_{ij}^{(l)(k)}(t) \; \middle| \; j \in \mathcal{N}_i(t) \right\} .
\end{equation}

We obtain each $\mathbf{y}_{ij}^{(l)(k)}(t)$ in this set by applying the $f_\mathrm{com}^{(l)}$ function of layer $l$ to the $(k-1)$-hop data at each neighbor $j$, summing, and then communicating the result:

\begin{equation}
\mathbf{y}_{ij}^{(l)(k)}(t) = \smashoperator[r]{\sum_{\mathbf{z} \in \mathcal{Y}_j^{(l)(k-1)}(t-1)}} \; f_\mathrm{com}^{(l)}\left(\mathbf{z}\right) .
\label{eq:yij}
\end{equation}

As a base case for this recursive definition, the 0-hop data $\mathcal{Y}_i^{(l)(0)}(t)$ is defined as $\mathbf{c}_i^{(l)}(t)$, the output of $f_\mathrm{input}$:

\begin{equation}
\mathcal{Y}_i^{(l)(0)}(t) = \left\{ \mathbf{c}_i^{(l)}(t) \right\} .
\label{eq:Y0}
\end{equation}

At each timestep, the input to the node update module of agent $i$ is given by the set of data from all neighborhoods $\mathcal{Y}_i^{(k)}(t)$ up to the user-defined maximum number of hops $K$:

\begin{equation}
\mathcal{Z}_i^{(l)}(t) = \left\{ \mathcal{Y}_i^{(l)(k)}(t) \; \middle| \; k \in [0..K] \right\} .
\label{eq:Zi}
\end{equation}

\subsection{Node Update Module}
\label{section:GNNnode}

The node update module handles all local computation after the required data has been received. It takes each $k$-hop neighborhood of aggregated data as an input, and produces an action output.

The node update module is comprised of two aggregation steps and three user-defined submodules. The first aggregation step combines the states from the neighbors $\mathcal{N}_i(t)$ of agent $i$ (summing along the same dimension as the aggregation operations in the message aggregation module). The second aggregation step combines data from the $K+1$ different neighborhoods. The three user-defined submodules are interspersed throughout the model in the spaces between the aggregation steps. We use $\mathbf{x}^{(l)}_i(t)$ to represent the output

\begin{equation}
\mathbf{x}^{(l)}_i(t) = f_\mathrm{final}^{(l)} \left( \smashoperator[r]{\sum_{k=0}^{K}} \left[ f_\mathrm{mid}^{(l)(k)} \left( \smashoperator[r]{\sum_{\mathbf{z} \in \mathcal{Y}_i^{(l)(k)}(t)}} f_\mathrm{pre}^{(l)(k)}\left(\mathbf{z}\right) \right) \right] \right)
\label{eq:GNN}
\end{equation}

where $f_\mathrm{pre}^{(l)(k)}$, $f_\mathrm{mid}^{(l)(k)}$, and $f_\mathrm{final}^{(l)}$ are all submodules from layer $l$ of the node update module. Optionally, $f_\mathrm{pre}^{(l)(k)}$ and $f_\mathrm{mid}^{(l)(k)}$ may include $K+1$ separate networks, each of which is applied to a specific neighborhood $k$.

In contrast to many other GNN architectures, this formulation does not immediately aggregate the data from the neighboring agents. Instead, it applies a transformation $f_\mathrm{pre}$ to each transmission before that data is aggregated together. Information is inevitably lost through the summation operation, but $f_\mathrm{pre}$ can ensure that the \textit{most important} information is preserved.

To illustrate the loss of information without $f_\mathrm{pre}$, let us examine an example where a GCN is used for formation control. The inputs to each node are the relative positions of its neighbors, and the output is a desired velocity. In this example, let us consider the local calculations of agent $0$ in the swarm, whose neighbors are agents $1$, $2$, and $3$. The relative positions of the neighbors are $p_1 = [-2,0]$, $p_2=[1,1]$, and $p_3=[1,-1]$. In a GCN, the first step after receiving a set of transmissions is to aggregate them together. So, we calculate the aggregated information from a $1$-hop neighborhood: $[-2,0]+[1,1]+[1,-1] = [0,0]$. This calculation only yields the center of mass, so clearly some important information has been lost. It does not tell agent $0$ how many neighbors it has, how far away they are, or in which direction they lie. As further demonstration, if the relative positions are multiplied by $-1$, then one would expect a different output (for example, the agent might move in the positive x direction instead of the negative x direction), but the GCN would receive the same input of $[0,0]$. It is impossible to map the same input to different outputs, so a GCN cannot produce the correct output in all cases. Therefore, there exists a set of policies which cannot be represented by a GCN, no matter what its network parameters are.

In contrast, the introduction of $f_\mathrm{pre}$ provides a strong theoretical guarantee. It allows the network to represent \textit{any} symmetric function of the incoming data. To prove that the node update module is a universal approximator, we must leverage the theorem that any multivariate function $f$ can be represented as a composition of univariate functions: $f(X) = \rho \left( \sum_{x \in X} \phi(x) \right)$ \cite{DeepSets}. In ModGNN, the inner function $\phi$ corresponds to $f_\mathrm{pre}$, the summation corresponds to the first aggregation in the node update module, and the outer function $\rho$ corresponds to $f_\mathrm{mid}$. This produces an intermediate compressed state for each $k$-hop neighborhood. Each neighborhood can then be composed together in a similar fashion, where $\phi$ corresponds to $f_\mathrm{mid}$ and $\rho$ corresponds to $f_\mathrm{final}$. In effect, the node update module can represent any function $g$ of the form $g(\{X_k \,|\, k \in [0..K]\}) = f_2(\{f_1(X_k) \;|\; k \in [0..K]\})$, where $X_k$ is the set of $k$-hop data from each neighbor, $f_1$ is a symmetric function of data from a set of neighbors within a given neighborhood, and $f_2$ is a symmetric function of data from different neighborhoods. It is important that the architecture limits $f_1$ to symmetric functions in order to maintain permutation equivariance. On the other hand, the architecture permits different $f_\mathrm{mid}$ functions to be applied to different $k$-hop neighborhoods, so it is possible for $f_2$ to also represent some non-symmetric functions (for example, it can represent a GCN, which applies a different weight matrix to each $k$-hop neighborhood).

\subsection{ModGNN Implementation}
\label{section:Our Implementation}

In order to evaluate ModGNN against baselines, we develop a naive, application-agnostic implementation. All three submodules within the node update module are implemented with MLPs to serve as universal approximators, so it is dubbed ModGNN-MLP. ModGNN-MLP is a single-layer GNN defined by the following implementations for each of the submodules:
\begin{itemize}
    \item{$f_\mathrm{input}$} : The identity function.
    \item{$f_\mathrm{com}$} : The incoming state is subtracted from the local state, which is equivalent to using the Laplacian as a GSO.
    \item{$f_\mathrm{pre}$} : A three layer MLP with an output dimension size of 10.
    \item{$f_\mathrm{mid}$} : A three layer MLP with an output dimension size of 10.
    \item{$f_\mathrm{final}$} : A three layer MLP where the output is the size of the agent's action space.
\end{itemize}

\section{EVALUATION}
\label{section:Evaluation}

We compare the expressivity of ModGNN-MLP and the baselines by evaluating their ability to represent a known expert algorithm. Once trained, we also evaluate each model's ability to generalize to new situations. If a model can extrapolate to environments which were not seen in the training process, then that demonstrates that it actually learned the underlying rules associated with the expert algorithm, as opposed to simply mapping inputs to outputs.

\subsection{Experiments}

\begin{figure}[]
      \centering
       \resizebox{1.0\linewidth}{!}{
         \includegraphics[]{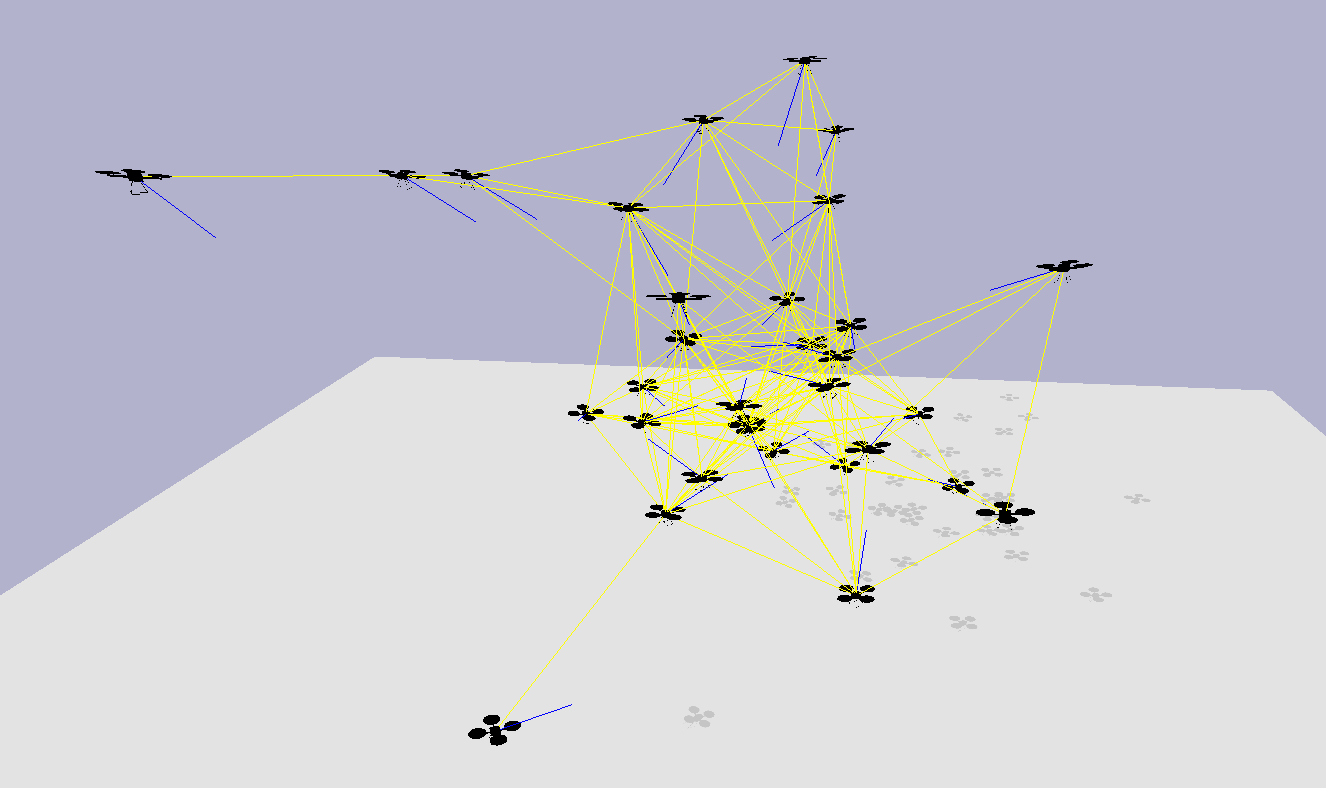}
        }
    \caption{A visualization of our PyBullet gym environment. The blue vectors represent the target velocities of each agent, and the yellow links represent edges in the communication graph.}
    \label{fig:Env}
\vspace*{-8pt}
\end{figure}

For our experiments, we focus on the flocking problem applied to a swarm of quadcopters \cite{TolstayaFlocking}. We create a gym environment in PyBullet \cite{Pybullet} (a physics simulator) and implement a PID velocity controller to translate actions given as target velocities into motor forces (Fig. \ref{fig:Env}).

Our chosen expert algorithm is Reynolds flocking, which combines elementary cohesion $\mathbf{c}_i(t)$, separation $\mathbf{s}_i(t)$, and alignment $\mathbf{a}_i(t)$ behaviors weighted by constants $C_c$, $C_s$, $C_a$ \cite{Reynolds}. In this formulation for the target velocity $\mathbf{u}_i(t)$ generated by Reynolds flocking, $\mathbf{p}_i(t)$ and $\mathbf{v}_i(t)$ represent the position and velocity of agent $i$:

\begin{equation}
\begin{split}
    \mathbf{c}_i(t) &= \sum_{j \in \mathcal{N}_i(t)} (\mathbf{p}_j(t)-\mathbf{p}_i(t)) \cdot  ||\mathbf{p}_j(t)-\mathbf{p}_i(t)|| \\
    \mathbf{s}_i(t) &= \sum_{j \in \mathcal{N}_i(t)} \frac{\mathbf{p}_i(t)-\mathbf{p}_j(t)}  {||\mathbf{p}_i(t)-\mathbf{p}_j(t)||^3} \\
    \mathbf{a}_i(t) &= \sum_{j \in \mathcal{N}_i(t)} (\mathbf{v}_j(t)-\mathbf{v}_i(t)) \cdot  ||\mathbf{v}_j(t)-\mathbf{v}_i(t)|| \\
    \mathbf{u}_i(t) &= C_\mathrm{c} \cdot \mathbf{c}_i(t) + C_\mathrm{s} \cdot \mathbf{s}_i(t) + C_\mathrm{a} \cdot \mathbf{a}_i(t) \;.
\end{split}
\label{eq:Reynolds}
\end{equation}

While the expert algorithm uses a fully connected communication graph, the models are trained to emulate that behavior with a communication range of $3.5$m. Using a simulation with 32 quadcopters, we train on a dataset of 1000 episodes (with randomized starting positions), each of length 1000 timesteps. At test time, we simulate 100 episodes with each model. In the swarm, one designated leader is controlled by generating a target velocity with integrated Gaussian noise. The raw observations of each agent $\mathbf{o}_i(t) \in \mathbb{R}^6$ are composed of a concatenation of their position and velocity.

\subsection{Models}
\label{section:Models}

We test our architecture against two variants of a GCN, a centralized network, and two \say{ablation models}, which are versions of ModGNN-MLP with various components stripped away:
\begin{itemize}
    \item{ModGNN-MLP} : An implementation of ModGNN (Section \ref{section:Our Implementation}).
    \item{ModGNN-MLP [$-f_\mathrm{pre}$]} : ModGNN-MLP with the $f_\mathrm{pre}$ submodule removed.
    \item{ModGNN-MLP [$-f_\mathrm{mid}$]} : ModGNN-MLP with the $f_\mathrm{mid}$ submodule removed.
    \item{GCN} : A standard GCN (Section \ref{section:GCN}). In the context of ModGNN, the GCN implements $f_\mathrm{mid}$ with a matrix multiplication and $f_\mathrm{final}$ with a point-wise activation function. The $f_\mathrm{pre}$ submodule is the identity function.
    \item{GCN [$+f_\mathrm{final}$]} : A variant of the GCN with an MLP in place of $f_\mathrm{final}$. This architecture has been used in previous work to represent more complex policies \cite{QingbiaoMAPF}.
    \item{Central} : A centralized MLP which takes the concatenated state space of all agents as an input, and produces the concatenated actions of all agents as an output. The architecture has 4 layers.
    
\end{itemize}

All models use the Laplacian GSO for $f_\mathrm{com}$ (except for the centralized model, which gets instantaneous access to all information). They are trained and evaluated in a centralized setting in order to benefit from the speed of vectorized operations. To ensure a fair comparison, all of the models are trained in parallel with the same batches of data, using the same hyperparameters (which are not optimized for any individual model).

\subsection{Results}
\label{section:Results}

In evaluation, we evaluate the models' performance not only over a validation dataset generated by an expert, but also over trajectories generated by their own policies. In these simulations, the same suite of randomized starting states is used for each model. Some of the evaluation environments have the same number of agents as in training, while others vary the number of agents in order to test the models' ability to generalize.

\begin{figure}[]
\begin{center}
\resizebox{!}{32pt}{
\begin{tabular}{|c||c|c|c|}
    \hline
    \multirow{2}{*}{Model} & \multicolumn{3}{c|}{Mean Squared Error} \\
    \cline{2-4}
    & $K=1$ & $K=2$ & $K=3$ \\ 
    \hline
    \hline
    \makecell[l]{ModGNN-MLP} & $\mathbf{0.037}$ & $\mathbf{0.033}$ & $\mathbf{0.032}$ \\
    \hline
    \makecell[l]{ModGNN-MLP [$-f_\mathrm{mid}$]} & $0.044$ & $0.037$ & $0.037$ \\
    \hline
    \makecell[l]{ModGNN-MLP [$-f_\mathrm{pre}$]} & $0.100$ & $0.078$ & $0.073$ \\
    \hline
    \makecell[l]{GCN} & $0.181$ & $0.149$ & $0.657$ \\
    \hline
    \makecell[l]{GCN [$+f_\mathrm{final}$]} & $0.111$ & $0.085$ & $0.086$ \\
    \hline
    \makecell[l]{Central} & $0.267$ & $0.267$ & $0.267$ \\
    \hline
\end{tabular}
}
\end{center}
\captionof{table}{\textbf{Validation Loss}. The mean squared error (relative to the expert). This is evaluated on trajectories generated by the expert in an environment with 32 agents. Note that the Central model gets instantaneous access to all agents' states, so changing the number of hops $K$ has no effect on it.}
\label{fig:ValidationLoss}
\vspace*{-8pt}
\end{figure}

\begin{figure}[]
\begin{center}
\resizebox{\linewidth}{!}{
\begin{tabular}{|l|c||c|c|c|c|}
\hline
\multicolumn{2}{|c||}{Model} & Error & Leader Dist & Cohesion & Separation \\
\hline
\hline
Expert & $K=\infty$ & $0.000$ & $3.55 \pm 3.96$ & $6.31 \pm 4.99$ & $0.87 \pm 0.96$ \\
\hline
\hline
ModGNN-MLP & $K=1$ & $\mathbf{0.197}$ & $\mathbf{3.76} \pm 1.91$ & $\mathbf{6.48} \pm 2.46$ & $0.86 \pm \mathbf{0.62}$ \\
\hline
ModGNN-MLP [$-f_\mathrm{mid}$] & $K=1$ & $0.208$ & $4.02 \pm 3.38$ & $6.89 \pm 4.61$ & $0.87 \pm 0.91$ \\
\hline
ModGNN-MLP [$-f_\mathrm{pre}$] & $K=1$ & $0.376$ & $4.13 \pm 2.10$ & $7.47 \pm 7.18$ & $0.78 \pm 1.30$ \\
\hline
GCN & $K=1$ & $0.363$ & $5.40 \pm 2.63$ & $8.56 \pm 2.11$ & $0.93 \pm 0.83$ \\
\hline
GCN [$+f_\mathrm{final}$] & $K=1$ & $0.373$ & $4.45 \pm 2.24$ & $7.85 \pm 5.91$ & $0.82 \pm 1.15$ \\
\hline
Central & $K=1$ & $0.395$ & $5.15 \pm 2.62$ & $9.69 \pm 3.96$ & $1.34 \pm 0.91$ \\
\hline
\hline
ModGNN-MLP & $K=2$ & $\mathbf{0.182}$ & $\mathbf{3.74} \pm 2.02$ & $\mathbf{6.52} \pm 2.52$ & $0.87 \pm \mathbf{0.63}$ \\
\hline
ModGNN-MLP [$-f_\mathrm{mid}$] & $K=2$ & $0.204$ & $3.78 \pm 2.06$ & $7.02 \pm 7.00$ & $0.89 \pm 1.34$ \\
\hline
ModGNN-MLP [$-f_\mathrm{pre}$] & $K=2$ & $0.285$ & $4.00 \pm 2.00$ & $8.35 \pm 11.0$ & $0.80 \pm 1.99$\\
\hline
GCN & $K=2$ & $0.318$ & $4.64 \pm 2.75$ & $8.82 \pm 3.27$ & $0.86 \pm 0.93$ \\
\hline
GCN [$+f_\mathrm{final}$] & $K=2$ & $0.314$ & $4.02 \pm 2.08$ & $7.65 \pm 5.65$ & $0.76 \pm 1.15$ \\
\hline
Central & $K=2$ & $0.395$ & $5.15 \pm 2.62$ & $9.69 \pm 3.96$ & $1.34 \pm 0.91$ \\
\hline
\hline
ModGNN-MLP & $K=3$ & $\mathbf{0.178}$ & $\mathbf{3.88} \pm 2.25$ & $\mathbf{6.68} \pm 2.77$ & $0.87 \pm \mathbf{0.64}$ \\
\hline
ModGNN-MLP [$-f_\mathrm{mid}$] & $K=3$ & $0.263$ & $4.70 \pm 6.79$ & $7.43 \pm 6.77$ & $0.89 \pm 1.35$ \\
\hline
ModGNN-MLP [$-f_\mathrm{pre}$] & $K=3$ & $0.278$ & $4.30 \pm 2.20$ & $7.72 \pm 5.90$ & $0.79 \pm 1.17$ \\
\hline
GCN & $K=3$ & $1.12$ & $7.73 \pm 10.6$ & $44.9 \pm 54.0$ & $1.93 \pm 7.44$ \\
\hline
GCN [$+f_\mathrm{final}$] & $K=3$ & $0.331$ & $5.16 \pm 2.98$ & $9.48 \pm 11.3$ & $0.85 \pm 2.27$ \\
\hline
Central & $K=3$ & $0.395$ & $5.15 \pm 2.62$ & $9.69 \pm 3.96$ & $1.34 \pm 0.91$ \\
\hline
\end{tabular}
}
\end{center}
\captionof{table}{\textbf{Model Evaluation}. The mean squared error (relative to the expert) and flocking metrics for each model in an environment with 32 agents. We report the metrics in the format [mean] $\pm$ [standard deviation]. Note that these metrics are computed on trajectories generated by the trained models (as opposed to the error in Table \ref{fig:ValidationLoss}, which is calculated on trajectories generated by the expert). The total error is higher because compounded local errors can cause the agents to enter states which would never be visited by the expert, and therefore are not present in the training dataset.}
\label{fig:ModelEvaluation}
\vspace*{-8pt}
\end{figure}

First, we evaluate the expressivity of each model by computing their loss over a validation dataset (Table \ref{fig:ValidationLoss}). The results show that the models can be grouped into four distinct levels of performance:
\begin{enumerate}[] 
    \item The models which exhibit the best ability to emulate the expert algorithm are ModGNN-MLP (with a mean error of $0.034$) and ModGNN-MLP [$-f_\mathrm{mid}$] (with a mean error of $0.039$). These are the only two architectures which utilize the $f_\mathrm{pre}$ submodule. It makes sense that $f_\mathrm{pre}$ is the most important place for a nonlinearity because it is possible to reformulate the Reynolds flocking algorithm in the context of the ModGNN framework such that all nonlinear operations are contained in $f_\mathrm{pre}$.
    \item The models with the second lowest validation loss are the ModGNN-MLP [$-f_\mathrm{pre}$] (with a mean error of $0.084$) and GCN [$+f_\mathrm{final}$] (with a mean error of $0.094$). These models do not have a nonlinear $f_\mathrm{pre}$ submodule, but they do have MLPs in other locations. They cannot precisely emulate Reynolds flocking because information is lost in the aggregation operation after $f_\mathrm{pre}$, but they can use their MLPs to (imperfectly) reconstruct and use the data.
    \item The third best model is the GCN (with a mean error of $0.165$, accounting for one outlier). It does not contain any MLPs, so it cannot capture the complexity of the nonlinear functions in Reynolds flocking. Instead, it regresses on the single linear layer model that reduces the error as much as possible. Interestingly, the GCN received a high error of $0.657$ for the case of $K=3$. It is unclear whether this is due to the $3$-hop data introducing noise which interferes with the other data, or if the model simply reached a local optimum during training.
    \item The Central model is the worst (with a mean error of $0.267$). Although it is given all of the necessary information and has an architecture that is complex enough to capture the intricacies of Reynolds flocking, it simply cannot generalize. The state space is far too large, so the probability of encountering a state that is close to one that it had seen in training is very low.  
\end{enumerate}

\begin{figure}[]
\begin{center}
\resizebox{!}{70pt}{
\begin{tabular}{|l|c||c|c|c|}
\hline
\multicolumn{2}{|c||}{\multirow{2}{*}{Model}} & \multicolumn{3}{c|}{Mean Squared Error} \\
\cline{3-5}
\multicolumn{2}{|c||}{} & $N=16$ & $N=32$ & $N=64$ \\ 
\hline
\hline
ModGNN-MLP & $K=1$ & $0.297$ & $\mathbf{0.197}$ & $\mathbf{0.291}$  \\
\hline
ModGNN-MLP [$-f_\mathrm{mid}$] & $K=1$ & $\mathbf{0.237}$ & $0.208$ & $0.394$ \\
\hline
ModGNN-MLP [$-f_\mathrm{pre}$] & $K=1$ & $0.393$ & $0.376$ & $0.297$ \\
\hline
GCN & $K=1$ & $0.392$ & $0.363$ & $0.323$ \\
\hline
GCN [$+f_\mathrm{final}$] & $K=1$ & $0.415$ & $0.373$ & $0.293$ \\
\hline
\hline
ModGNN-MLP & $K=2$ & $\mathbf{0.271}$ & $\mathbf{0.182}$ & $\mathbf{0.234}$  \\
\hline
ModGNN-MLP [$-f_\mathrm{mid}$] & $K=2$ & $0.395$ & $0.204$ & $0.270$ \\
\hline
ModGNN-MLP [$-f_\mathrm{pre}$] & $K=2$ & $0.367$ & $0.285$ & $0.265$ \\
\hline
GCN & $K=2$ & $0.426$ & $0.318$ & $0.437$ \\
\hline
GCN [$+f_\mathrm{final}$] & $K=2$ & $0.492$ & $0.314$ & $0.400$ \\
\hline
\hline
ModGNN-MLP & $K=3$ & $\mathbf{0.277}$ & $\mathbf{0.178}$ & $\mathbf{0.287}$ \\
\hline
ModGNN-MLP [$-f_\mathrm{mid}$] & $K=3$ & $0.290$ & $0.263$ & $0.358$ \\
\hline
ModGNN-MLP [$-f_\mathrm{pre}$] & $K=3$ & $0.326$ & $0.278$ & $0.468$ \\
\hline
GCN & $K=3$ & $1.09$ & $1.12$ & $1.18$ \\
\hline
GCN [$+f_\mathrm{final}$] & $K=3$ & $0.525$ & $0.331$ & $0.417$ \\
\hline
\end{tabular}
}
\end{center}
\captionof{table}{\textbf{Generalization Tests}. The mean squared error for each model in different environments. The models are trained with 32 agents, and they are tested with 16, 32, and 64 agents.}
\label{fig:GeneralizationTests}
\vspace*{-8pt}
\end{figure}

Next, we use the models' learned policies in simulation (Table \ref{fig:ModelEvaluation}). The resulting trajectories are evaluated with the following metrics:
\begin{itemize}
    \item Error : The mean squared error between the model output and the action produced by the expert algorithm.
    \item Leader Distance : The distance from the leader to the center of mass of the rest of the swarm. This is a measure of the swarm's responsiveness to decisions made by the leader.
    \item Cohesion : The diameter of the smallest sphere that can contain all of the agents. It is desirable for the cohesion value to be as small as possible (until it starts conflicting with separation), because that indicates that the agents are sticking together in a swarm.
    \item Separation : The distance between each pair of agents. The separation should not be too low (resulting in collisions) or too high (resulting in poor cohesion). Most importantly, the separation metric should have a low standard deviation. A consistent separation between agents implies an efficient formation with regular spacing.
\end{itemize}

The results show that ModGNN-MLP performs better than the baselines and ablation models across all metrics for all values of $K$ that we tested (Table \ref{fig:ModelEvaluation}). The mean squared error in simulation is always at least $42\%$ better than the best baseline for all values of $K$. In the standard deviation of the separation---perhaps the most telling metric---ModGNN-MLP outperforms both variants of the GCN by achieving a value that is $25\%$ smaller for $K=1$, $33\%$ smaller for $K=2$, and $72\%$ smaller for $K=3$. This improvement in performance with increasing $K$ is also reflected in the mean squared error of ModGNN-MLP, but it is not replicated by the baselines. This indicates that unlike the other architectures, ModGNN-MLP is able to utilize multi-hop information to improve its predictions.

The last set of experiments that we perform are a series of generalization tests (Table \ref{fig:GeneralizationTests}). We apply the models (which have been trained with 32 agents) to new environments with 16 or 64 agents, thereby evaluating their ability to generalize.

Again, the results indicate that ModGNN-MLP consistently outperforms the baselines, demonstrating an ability to produce accurate results even when the agents are presented with a varying number of neighbors. In fact, the loss of ModGNN-MLP in the new environments ($N=16$ and $N=64$) is lower than the loss of the baselines in the same environment on which they were trained ($N=32$). The only test case where ModGNN-MLP is not the best is for $K=1$, $N=16$, where it is beaten by ModGNN [$-f_\mathrm{mid}$]. This is not surprising, as ModGNN [$-f_\mathrm{mid}$] establishes itself as a close runner-up to ModGNN-MLP in the previous tests (Table \ref{fig:ValidationLoss}, \ref{fig:ModelEvaluation}). As the only other model which includes the $f_\mathrm{pre}$ submodule, it has the ability to represent the ground truth function which the expert applies to the incoming data from each neighbor. 

\section{DISCUSSION}
\label{section:Discussion}

These results demonstrate that a naive implementation of the ModGNN framework is able to approximate an expert algorithm more accurately than the baselines. Furthermore, the ModGNN implementation shows an improved ability to generalize to previously unseen environments, which indicates that it is able to learn the underlying rules in the expert algorithm. The results also provide experimental evidence to back up the theoretical advantages of including the $f_\mathrm{pre}$ submodule in a GNN architecture.

Of course, these results only reflect the benefits of our framework in a single case study. In the future, it would be worthwhile to apply the same comparison between models to more applications. In doing so, we could determine if $f_\mathrm{pre}$ is always the most important submodule, or if the importance of each individual submodule varies depending on the application.

In this paper, we focused on architectural changes to submodules within the node update module. It would also be interesting to evaluate the effect of changes within the message aggregation module. For example, the implementation of $f_\mathrm{com}$ determines which of three categories a GNN falls under: standard \cite{NodeClassification}, attention-based \cite{GraphAttention}, or message-passing \cite{MessagePassing}. One could evaluate a novel implementation of $f_\mathrm{com}$ against baselines from these three categories.

Future research could also delve into the effects of changing other hyperparameters: Does the importance of using nonlinear models in the submodules change when the GNN has more layers? What implementations of ModGNN are most robust to changes in the communication range? The answers to these questions could inform the development of new implementations of ModGNN.

\section{CONCLUSION}

In this paper, we proposed ModGNN, a new framework for GNNs in multi-agent problems. We used ModGNN to implement our own novel GNN, as well as baseline architectures for comparison. We demonstrated that the addition of extra nonlinear submodules in a GNN can greatly boost its performance. Furthermore, we showed that a naive implementation of ModGNN posesses more expressive power than a GCN, and demonstrates an improved ability generalize to new scenarios.

\vspace{10pt}
\section{Acknowledgements}

\footnotesize{This work is partially supported by Nokia Bell Labs through their donation for the Centre of Mobile, Wearable Systems and Augmented Intelligence to the University of Cambridge.
A. Prorok was supported by the Engineering and Physical Sciences Research Council (grant EP/S015493/1) and ERC Project 949940 (gAIa).
We also thank Jan Blumenkamp for testing and providing feedback about the ModGNN Python library.}


\addtolength{\textheight}{-12cm}




\bibliographystyle{Styles/IEEEtran}
\bibliography{ref}

\begin{thebibliography}{10}
\providecommand{\url}[1]{#1}
\csname url@rmstyle\endcsname
\providecommand{\newblock}{\relax}
\providecommand{\bibinfo}[2]{#2}
\providecommand\BIBentrySTDinterwordspacing{\spaceskip=0pt\relax}
\providecommand\BIBentryALTinterwordstretchfactor{4}
\providecommand\BIBentryALTinterwordspacing{\spaceskip=\fontdimen2\font plus
\BIBentryALTinterwordstretchfactor\fontdimen3\font minus
  \fontdimen4\font\relax}
\providecommand\BIBforeignlanguage[2]{{%
\expandafter\ifx\csname l@#1\endcsname\relax
\typeout{** WARNING: IEEEtran.bst: No hyphenation pattern has been}%
\typeout{** loaded for the language `#1'. Using the pattern for}%
\typeout{** the default language instead.}%
\else
\language=\csname l@#1\endcsname
\fi
#2}}

\bibitem{OriginalGNN}
F.~{Scarselli}, M.~{Gori}, A.~C. {Tsoi}, M.~{Hagenbuchner}, and
  G.~{Monfardini}, ``The graph neural network model,'' \emph{IEEE Transactions
  on Neural Networks}, vol.~20, no.~1, pp. 61--80, 2009.

\bibitem{GraphNetworks}
P.~W. Battaglia, J.~B. Hamrick, V.~Bapst, A.~Sanchez-Gonzalez, V.~Zambaldi,
  M.~Malinowski, A.~Tacchetti, D.~Raposo, A.~Santoro, R.~Faulkner,
  \emph{et~al.}, ``Relational inductive biases, deep learning, and graph
  networks,'' \emph{arXiv preprint arXiv:1806.01261}, 2018.

\bibitem{Ribeiro}
L.~Ruiz, F.~Gama, and A.~Ribeiro, ``Graph neural networks: Architectures,
  stability, and transferability,'' \emph{Proceedings of the IEEE}, 2021.

\bibitem{NodeClassification}
T.~N. Kipf and M.~Welling, ``Semi-supervised classification with graph
  convolutional networks,'' \emph{International Conference on Learning
  Representations (ICLR)}, 2017.

\bibitem{GraphAttention}
P.~Velickovic, G.~Cucurull, A.~Casanova, A.~Romero, P.~Li{\`o}, and Y.~Bengio,
  ``Graph attention networks,'' \emph{International Conference on Learning
  Representations (ICLR)}, 2018.

\bibitem{GraphSAGE}
W.~Hamilton, Z.~Ying, and J.~Leskovec, ``Inductive representation learning on
  large graphs,'' in \emph{Advances in Neural Information Processing Systems},
  2017, pp. 1024--1034.

\bibitem{GraphClassification}
D.~Bacciu, F.~Errica, and A.~Micheli, ``Contextual graph markov model: A deep
  and generative approach to graph processing,'' in \emph{International
  Conference on Machine Learning}.\hskip 1em plus 0.5em minus 0.4em\relax PMLR,
  2018, pp. 294--303.

\bibitem{LinkPrediction}
M.~Zhang and Y.~Chen, ``Link prediction based on graph neural networks,'' in
  \emph{Proceedings of the 32nd International Conference on Neural Information
  Processing Systems}, 2018, pp. 5171--5181.

\bibitem{ArbaazLabelled}
A.~Khan, E.~Tolstaya, A.~Ribeiro, and V.~Kumar, ``Graph policy gradients for
  large scale robot control,'' in \emph{Conference on Robot Learning}.\hskip
  1em plus 0.5em minus 0.4em\relax PMLR, 2020, pp. 823--834.

\bibitem{ArbaazUnlabelled}
A.~Khan, V.~Kumar, and A.~Ribeiro, ``Graph policy gradients for large scale
  unlabeled motion planning with constraints,'' \emph{ArXiv}, vol.
  abs/1909.10704, 2019.

\bibitem{TolstayaFlocking}
E.~Tolstaya, F.~Gama, J.~Paulos, G.~Pappas, V.~Kumar, and A.~Ribeiro,
  ``Learning decentralized controllers for robot swarms with graph neural
  networks,'' in \emph{Conference on Robot Learning}.\hskip 1em plus 0.5em
  minus 0.4em\relax PMLR, 2020, pp. 671--682.

\bibitem{QingbiaoMAPF}
Q.~Li, F.~Gama, A.~Ribeiro, and A.~Prorok, ``Graph neural networks for
  decentralized multi-robot path planning,'' in \emph{IEEE/RSJ International
  Conference on Intelligent Robots and Systems}, 2020.

\bibitem{GNNCoverage}
E.~Tolstaya, J.~Paulos, V.~Kumar, and A.~Ribeiro, ``Multi-robot coverage and
  exploration using spatial graph neural networks,'' \emph{arXiv preprint
  arXiv:2011.01119}, 2020.

\bibitem{ProrokWorkshop}
A.~Prorok, ``Graph neural networks for learning robot team coordination,'' in
  \emph{ICML workshop on Federated AI for Robotics, arXiv preprint
  arXiv:1805.03737}, 2018.

\bibitem{GamaGCN}
F.~Gama, A.~G. Marques, G.~Leus, and A.~Ribeiro, ``Convolutional neural network
  architectures for signals supported on graphs,'' \emph{IEEE Transactions on
  Signal Processing}, vol.~67, no.~4, pp. 1034--1049, 2018.

\bibitem{DeepSets}
M.~Zaheer, S.~Kottur, S.~Ravanbhakhsh, B.~P{\'o}czos, R.~Salakhutdinov, and
  A.~J. Smola, ``Deep sets,'' in \emph{Proceedings of the 31st International
  Conference on Neural Information Processing Systems}, 2017, pp. 3394--3404.

\bibitem{Pybullet}
E.~Coumans and Y.~Bai, ``Pybullet, a python module for physics simulation in
  robotics, games and machine learning,'' 2017.

\bibitem{Reynolds}
C.~W. Reynolds, ``Flocks, herds and schools: A distributed behavioral model,''
  in \emph{Proceedings of the 14th Annual Conference on Computer Graphics and
  Interactive Techniques (SIGGRAPH)}, 1987, pp. 25--34.

\bibitem{MessagePassing}
J.~Gilmer, S.~S. Schoenholz, P.~F. Riley, O.~Vinyals, and G.~E. Dahl, ``Neural
  message passing for quantum chemistry,'' in \emph{International Conference on
  Machine Learning}.\hskip 1em plus 0.5em minus 0.4em\relax PMLR, 2017, pp.
  1263--1272.

\end{thebibliography}

\end{document}